\documentclass[letterpaper, 10 pt, conference]{ieeeconf}  

\IEEEoverridecommandlockouts                              
                                                          
\usepackage{cite}
\usepackage{amsmath,amssymb,amsfonts}
\usepackage{algorithmic}
\usepackage{graphicx}
\usepackage{textcomp}
\usepackage{xcolor}
\usepackage{booktabs}
\usepackage{microtype}
\usepackage{array}
\usepackage{booktabs}
\usepackage{textcomp}
\usepackage{float}
\usepackage{censor}
\usepackage[hyphens]{url}
\usepackage{hyperref}
\usepackage{breakurl}
\let\labelindent\relax
\usepackage[inline]{enumitem}
\usepackage{subcaption}
\usepackage[font=small,skip=2pt]{caption}
\def\BibTeX{{\rm B\kern-.05em{\sc i\kern-.025em b}\kern-.08em
    T\kern-.1667em\lower.7ex\hbox{E}\kern-.125emX}}
\begin{document}
\setlength{\textfloatsep}{6pt}
\setlength{\dbltextfloatsep}{6pt}
\setlength{\floatsep}{6pt}

\title{\LARGE \bf Examining the Difference in Human Behavior Between Virtual and Real-World Human-Robot Teaming

\thanks{$^{1}$M. AbuHijleh, S. Dallas, A. Getachew, W. Louie, and A. Macklem-Zabel are with the Department of Electrical and Computer Engineering, Oakland University, Rochester, MI 48309, United States
{\tt\small{abuhijleh, srdallas, absalatgetachew, macklemzabel, louie}@oakland.edu}}
\thanks{$^{2}$  D. Zytko is with the College of Innovation \& Technology, University of Michigan-Flint, Flint, MI 48502, United States
{\tt\small dzytko@umich.edu}}
\thanks{$^{3}$ M. Brudnak is with U.S. Army DEVCOM Ground Vehicle Systems Center (GVSC), United States
{\tt\small mark.j.brudnak.civ@army.mil}}
\thanks{DISTRIBUTION STATEMENT A. Approved for public release; distribution is unlimited. OPSEC10425.}
}

\author{Sean Dallas$^{1}$, Absalat Getachew$^{1}$, Motaz AbuHijleh$^{1}$, Andrea Macklem-Zabel$^{1}$, \\ Douglas Zytko$^{2}$, Mark Brudnak$^{3}$, and Wing-Yue Geoffrey Louie$^{1}$}

\maketitle
\begin{abstract}

Prototyping and evaluating human-robot teaming (HRT) scenarios in the real-world is costly. Virtual simulation of HRT scenarios has been adopted as an alternative to conducting user studies in the real-world to investigate user perceptions, behaviors, and performance during human-robot interactions. The consistency of human behavior between the real and virtual-worlds is integral to the validity of utilizing such virtual experimentation. This paper presents a user study examining the difference in human behavior between an HRT scenario conducted in a virtual vs real environment. We employed mixed-methods to examine team performance and human factors during the HRT. Our quantitative results showed a significant difference in workload between the two modalities. Our qualitative analysis expanded on the quantitative results and found differences in participants' strategy, their mental model of robots, and the type of trust they had for robots between modalities.

\end{abstract}
\vspace{4pt}


\section{Introduction}

As robotic technology continues to advance, robots are increasingly being integrated into teams where they work alongside humans to achieve shared goals. Robots are used in these teams to mitigate human risks, improve situational awareness (SA), and harness their computational resources to complete tasks more efficiently \cite{wolf2023and, robinson2024human, gervits2020toward, walker2024cyber}. This paradigm is referred to as Human-Robot Teaming (HRT). HRT represents a distinct and increasingly important area of human-robot interaction (HRI), where multiple humans and robots collaborate to achieve shared objectives \cite{mcneese2018teaming}. 

Despite the potential of HRT, progress is often slow as it is costly and logistically challenging to prototype these scenarios. Virtual simulation potentially addresses these challenges by supporting rapid iteration and testing of robots in environments that would be difficult or impractical to replicate physically, such as hazardous disaster zones or crowded urban areas \cite{wang2015intelligent}. Furthermore, virtual simulations offer safety, control, and greater cost-effectiveness.

Despite the advantages of virtually prototyping HRT, the existing literature for one-on-one HRIs has frequently identified discrepancies in human perception between interactions in the virtual-world (VW) and real-world (RW)  \cite{li2019comparing, wijnen2020performing, weistroffer2014assessing}. These differences raise questions about the accuracy and transferability of insights gained from virtual HRI. Unlike dyadic HRI, which often emphasizes supervisory control or tool use, HRT introduces core elements of team cognition such as shared mental models and mutual predictability \cite{gervits2020toward}. In contrast to dyadic HRI, where attention can be devoted to a single agent, HRT necessitates the distribution of attention across multiple teammates. This is because team members are often not co-located and, consequently, critical information about the robot state as well as actions is not directly observable. Therefore, a task which taxes trust, SA, and workload while working in a human-robot team is necessary to investigate the effects modalities have on human factors and behaviors during HRT \cite{hopko2022human}. 

This research investigates the potential differences in human behavior and performance between VW and RW HRT scenarios. We conducted a mixed-methods between-subjects study where participants engaged in an HRT security task either in the VW or RW. Our approach incorporated quantitative assessments of human perception (via questionnaires) and task performance (via performance metrics), along with qualitative insights from think-aloud protocols and post-scenario interviews. The independent variable we controlled was the modality in which participants experienced the HRT scenario. To reduce potential threats to internal validity stemming from modality differences our VW was designed to replicate the RW scenario as closely as possible \cite{campbell2015experimental}. This allowed us to compare key HRT factors such as human trust, SA, workload, and strategy in a controlled and interactive setting. 

\section{Related Works} 
\subsection{Human Factors}

HRI studies rely on human factor measures to model and understand human behavior within HRIs. These human factor measures aim to capture human internal states during HRIs and can be integrated into human-in-the-loop control systems \cite{abdulazeem2023human}. Commonly studied human factors include trust, SA, workload, and related constructs \cite{hopko2022human}. 

\subsubsection{Trust}

 The cost-efficiency of VW experiments allows more frequent and varied interactions with robots. Studies like Adami et al. show that this increased exposure makes VWs a more effective tool for building trust, simply because it provides more opportunities for the user to see what the robot can do \cite{adami2022impact}. When the frequency of exposure and experiences with robots is constant, prior studies comparing trust across virtual and real HRI suggest that trust differences are not solely determined by interaction modality, but are shaped by how the interaction environment mediates experience and understanding of robot behavior. For example, Plomin et al. found no significant differences in perceived or behavioral trust between VW and RW interactions with a robot \cite{plomin2023virtual}. In contrast, Wijnen et al. found that participants initially had less trust-consistent behaviors compared to RW interactions when participants did not perceive robots as social others. However, trust behaviors eventually converged between modalities after exposure \cite{wijnen2020performing}. Together, these findings indicate that VWs can approximate RW trust with enough time, but trust may diverge as experience and perceptions of robots between the modalities differ. As a result it remains an open question whether trust develops similarly across virtual and physical settings in multi-robot HRT tasks, where trust must be distributed, calibrated, and maintained across multiple autonomous agents.

\subsubsection{Situational Awareness}

Evidence from physical and virtual HRI experiments suggests that SA is shaped by modality constraints. While RW settings are often restricted by safety protocols and hardware costs, VWs allow users to freely experience edge case scenarios and high risk situations. For instance, Adami et al. found that VW participants developed significantly higher SA than those in the RW, a gap attributed to the enhanced experiences gained in virtual reality (VR) \cite{adami2022impact}. This is due to the virtual modality enabling exposure to robotic failures that would be too dangerous to replicate physically and, as a result, participants adopted more robust monitoring strategies. Conversely, Horsch et al. \cite{horsch2013comparing} found no significant difference in SA when search-and-rescue tasks were kept identical across both conditions. Collectively, these findings suggest that SA is not a product of the medium itself, but of the breadth of interaction it permits; when the range of possible scenarios is equalized across modalities, cognitive outcomes converge.

\subsubsection{Workload}

HRI papers that compare operator workload within different modalities have found disparities in how physical and mental consequences differ between modalities \cite{tsoi2024influence, sakib2021physiological, adami2022impact}. For example, Sakib et al. \cite{sakib2021physiological} provided statistical evidence that a VW task led to higher mental workload than in a physical-world version of the same task. The authors found that while VR imposed higher cognitive demand, the physical-world led to different physiological responses including elevated skin conductance and temperature caused by environmental factors such as wind and humidity. However, this differing workload does not always lean towards VW interactions having higher mental workload. For instance, Adami et al. \cite{adami2022impact} found that VR based training did not reveal a significant difference in workload between modalities. The authors noted this was driven by limited physical realism and a lack of physical consequence in the VW. Together, these findings suggest that the relationship between interaction modality and operator workload is inconsistent, and may depend on the degree of physical realism and environmental fidelity offered by the VW. 

\subsection{Human Behavior in Real and Virtual Modalities} 

Beyond standard human factors like Trust, Workload, and SA, several HRI studies have noted that human behavior toward robots can differ significantly between modalities. 

For example, Li et al. \cite{li2019comparing} investigated human-robot proxemics and found that participants allowed closer proximity in the real world as well as reported a stronger sense of presence compared to VR. Furthermore, Weistroffer et al. \cite{weistroffer2014assessing} explored co-presence on industrial assembly lines and found that participants in the real world completed more tasks as well as felt a greater need to adapt their behavior compared to those in a virtual CAVE environment. Although Plomin et al. \cite{plomin2023virtual} found no statistically significant differences in reported trust or engagement between VR, screen-based, and RW conditions, they did observe that proxemic differences converged over time. Ultimately, these findings suggest that interactive VW HRI can illicit similar behavioral outcomes after acclimatization.

\begin{figure}
    \centering
    \vspace{6pt}
    \includegraphics[width=1\linewidth]{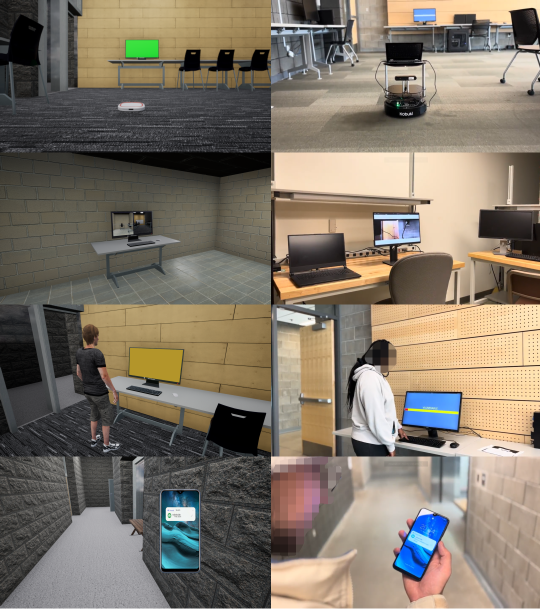}
    \caption{The left side is VW, right side is RW condition. The top row is of the SR2, the 2nd row is of the MR, the 3rd shows a hacker, and the bottom row is in building B showing a notification.}
    \label{fig:comparisonIMG}
\end{figure}

\section{Methods}
We designed an HRT security scenario to investigate the effect interaction modality has on human factors in a human-robot team and conducted a between-subjects user study with modality (screen-based VW vs. RW) as the independent variable. All participants experienced the same security scenario; the only difference across conditions was the modality (VW or RW). Participants were randomly assigned to one condition. We investigated differences across key human factors to examine the strengths and limitations of VW-based HRT experimentation. Photos of both experimental modalities are shown in Figure \ref{fig:comparisonIMG}. To the best of our knowledge, our study is the first to compare human behavior and perception in an HRT with multiple robot teammates across real and virtual worlds. The procedures for collecting data and running the study were reviewed and approved by the University Institutional Review Board, IRB \#FY2023-41. 

\subsection{Hypotheses}


We developed four hypotheses based on existing literature:

\begin{itemize}

    \item H1: Participants will exhibit comparable levels of trust in the VW and RW. While initial trust may differ, we hypothesize it will converge over time after sufficient exposure to robotic teammates, consistent with \cite{plomin2023virtual}.

    \item H2: Participants will have higher mental workload in the VW and a higher physical workload in the RW.

    \item H3: Participants will demonstrate equivalent levels of SA in both the VW and RW conditions.

    \item H4: Participants will have higher levels of presence in the RW compared to VW, aligning with \cite{li2019comparing}.
\end{itemize}

\subsection{Participants}

The inclusion criteria were: 1) 18-64 years old, 2) fully vaccinated for COVID-19, 3) able to move 1 meter per second while walking, and 4) fluent in English. We excluded participants with hearing impairments because participants needed to listen for audio queues during experiments. A total of 33 participants were recruited. Three participants were removed due to technical issues during the experiment, and 2 were removed for not following instructions. Our final sample included 28 participants (20 male, 8 female). Participants were randomly assigned and balanced across the RW and VW conditions. The VW group contained 9 males and 5 females (age $M = 22.57, std = 3.33$), the RW group contained 11 males and 3 females (age $M = 23.57, std = 5.19$). A monetary incentive was used to encourage participants to try their best and advertised as a range of \$10-\$15 based on their performance.

\subsection{Security Scenario}
\begin{figure}
    \centering
        \vspace{6pt}
    \includegraphics[width=1\linewidth]{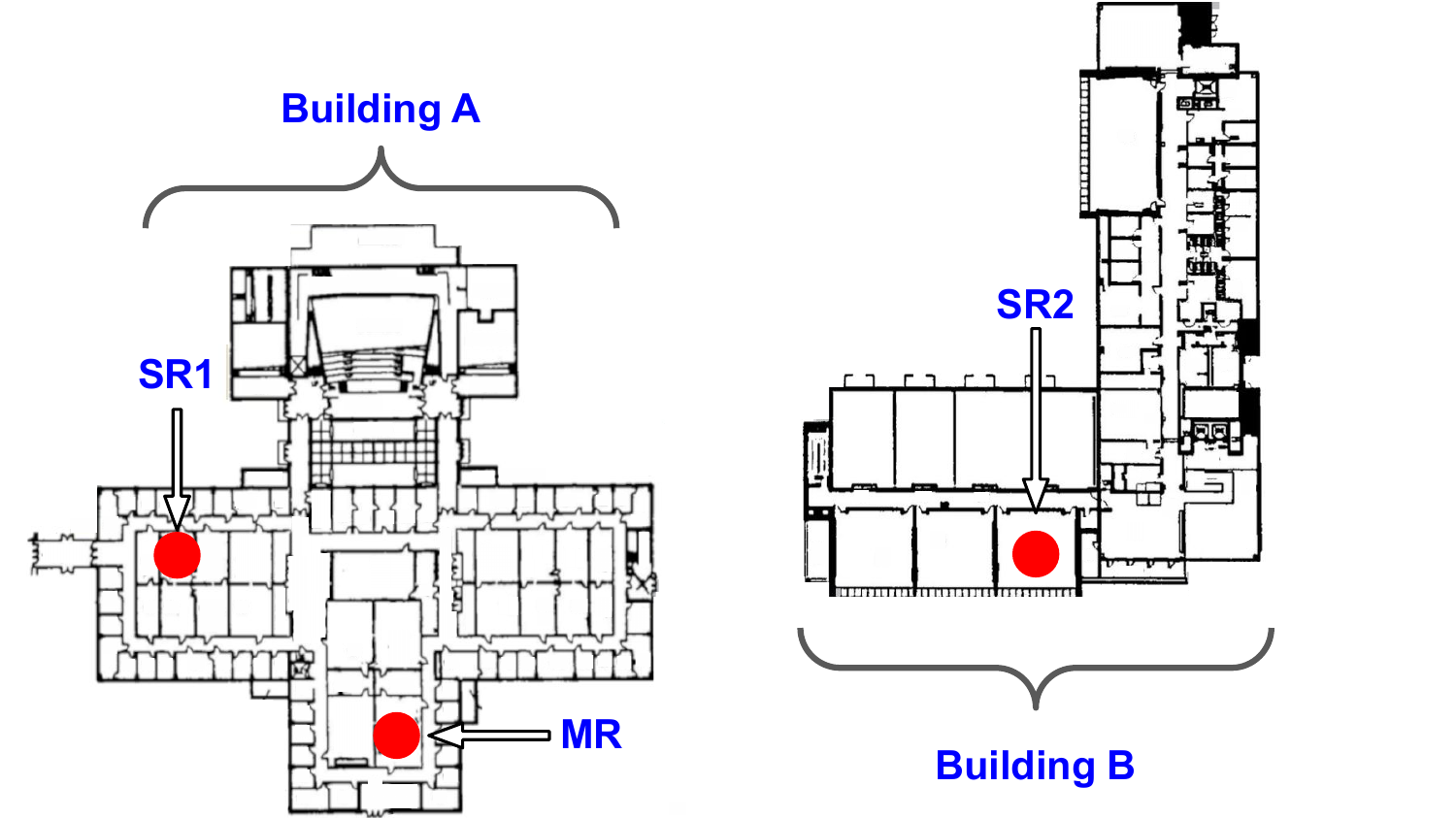}
    \caption{Map of room locations on campus.}
    \label{fig:Ou-map}
\end{figure}

Participants were informed that they were assigned a security role and needed to maintain the security and temperature of servers located within buildings on a university campus. The layout consisted of two buildings (A and B) with three key rooms between them. Building A had two rooms: a monitoring room (MR) and a server room (SR). Building B had one SR. The location of these rooms can be seen in Figure \ref{fig:Ou-map}. Each room housed a security robot, a server if it was a SR, and a monitoring PC if it was the MR. The rooms and buildings were chosen for their distance apart from each other because they required a non-trivial amount of walking to traverse between them. The rationale was to tax participants' SA and planning skills to navigate the spaces. Overall, the participants' two main objectives were to (1) maintain the security of the servers located in buildings A and B and (2) ensure the servers did not overheat. 

\subsubsection{Server Security}
During the scenario, hackers tried to compromise servers by pressing a keyboard button for 30 seconds. If a participant moved within 1m of a hacker, the process was interrupted and the server was left in a "vulnerable" state. However, the server became "hacked" if the hacker was not stopped. Participants had to manually re-secure servers in either state but the time required differed. Re-securing a vulnerable server took 15 seconds, but a fully hacked server required a longer 30-second button press. Participants were warned that their overall score would decrease any time a server remained in the hacked state. 

\subsubsection{Server Temperature}
Each server's temperature starts at 20°C and had seed based pseudo-random fluctuations but overall rose over time. Servers required periodic venting to maintain a healthy server temperature of 20-80°C. Venting the server when it is at a healthy temperature could be done at either the MR PC via a button press or a server within the SRs. Additionally, the participants were told to prevent the server from reaching 100°C as it would overheat and negatively affect their score. If the temperature of a server rose above 80°C, that server could no longer be vented from the MR, but instead must be vented from the SR. This dynamic forced participants to think about planning their vents and their routes to ensure the servers stayed within a safe temperature.

\subsubsection{Security Technology Description}
Within our study several key different technologies are incorporated to make our HRT scenario: 1) a security robot which works with participants to patrol SRs, 2) servers to keep secure and healthy, and 3) a phone that acts as a communication device between robots and the human.

The security robot located in each SR was a Turtlebot2 \cite{turtlebot2}. The robot platform was equipped with an RGB camera and a depth camera for navigation using SLAM. The Turtlebot2 autonomously navigated a pre-determined patrol route, relayed video feeds, and sent alerts when hackers were detected.  

Each server within the scenario was a PC with an interface displaying its temperature, security level, and a button to vent the server. The MR PC displayed video feeds from both security robots in each SR, the temperatures of each SRs server, and a button to vent the servers. 

Lastly, the participant had a mobile phone that was used for communication and receiving alerts. Alerts were sent by the patrolling Turtlebot2s when hackers were detected and when server temperatures reached 80°C and 100°C.

The design of this scenario was intentional in taxing trust, SA, and workload within an HRT scenario. Trust was engaged through the participant's reliance on the patrolling robots to help maintain server security and required participants to develop calibrated expectations of robot behavior into their HRT strategy. SA was challenged through the spatial distribution of critical information across servers and robots that were not co-located. The distribution of information required participants to integrate status updates from multiple sources that were not always directly observable. The dual-objective structure (balancing security and temperature management) imposed cognitive workload by demanding continuous attention switching, route planning, and prioritization of objectives across spatially separated tasks. Additionally, the temperature threshold mechanic (preventing remote venting above 80°C) created time pressure that compounded workload as participants had to anticipate future states and coordinate preemptively. Together, these elements created a realistic HRT scenario in which participants had to maintain awareness of a dynamic environment, manage competing demands under cognitive load, and coordinate with autonomous teammates. 

\subsection{Virtual Simulation}
In order to conduct a comparison between the VW and RW HRT experiences, we developed a digital twin that replicates the RW security scenario using Unreal Engine (Figure \ref{fig:comparisonIMG}). We ensured interactions between participants, robots, and the environment were as close to one-to-one as possible. The VW environment was modeled using building schematics to ensure the floor plans and overall size of the buildings are identical to the RW. The robots used in the simulation  modeled the RW security robot and their speed and patrol patterns were set identical to those in the RW. The participants VW walking speed was set to match their RW walking speed. Notifications were handled inside the simulation and were displayed using a toggleable 2D UI element that mimics the UI in the phone used in the RW. 


\subsection{Procedure}

Firstly, informed consent was obtained from each participant. Afterwards, participants' walking speed was measured by averaging three walks between two pieces of tape placed 6 meters apart. This measured speed was used to set the speed of participants embodiment in the VW condition. An explanation of the HRT scenario alongside a tour of the buildings was conducted in each condition. Participant questions were answered after the explanation and during the tour. For the RW condition, they were asked not to run and avoid any physical contact with the hackers. After the tour, a demographic survey and pre-trial Trust Perception Scale-HRI questionnaire were administered to establish a trust baseline. Participants in the RW condition were then given a phone for communication and alerts during the trials. After, a forward facing camera was strapped to their chest, to verify at a later time that participants stopped moving during their pauses. For both conditions, the participants were given headphones equipped with a microphone for communication and sound alerts. 

Participants then completed three 15-minute trials (excluding pauses), all starting from the monitoring room. The three trials were designated Trial A, B, and C. All participants regardless of condition experienced one of two orders for the preliminary trials: (A,B) or (B,A). The trials were counterbalanced across participants and followed by a fixed final test trial (C) to control for order effects. Participants would answer Trust Perception Scale-HRI, NASA-TLX, and SUS questionnaires after each trial. Lastly, participants were interviewed once all trials were finished. The procedure took two hours to finish.

\subsection{Mixed Methods Data Collection and Analysis}
A convergent mixed-methods design \cite{creswell2017designing, guetterman2015integrating} was employed, where quantitative and qualitative data were collected and analyzed concurrently yet separately as primary methods.

\subsection{Quantitative Data Collection Metrics}

We measured participants' trust, SA, workload, presence, and performance.

\subsubsection{Performance}

Performance was assessed using two scores 1) the total time the rooms were under hacker control and 2) the total time a server remained overheated.

\subsubsection{Trust}

Trust was measured using the reduced version of the 40-item Trust Perception Scale-HRI \cite{mittu_measuring_2016}, using 14 out of the 40 items that were most relevant to our scenario. The participant's trust score was calculated by averaging the values of the items.

\subsubsection{Situational awareness}

We measured SA using the Situational Awareness Global Assessment Technique (SAGAT) \cite{endsley_situation_1988}. The SAGAT questions asked were designed to assess participants' awareness and understanding of the current state of the environment and their ability to predict future conditions based on the information they had. At four predetermined pauses per trial, researchers verbally asked two of six questions about the current environment and its future state. The timing and ordering of questions were different for each of the three trials but the same for all participants. In the RW, a phone notification and verbal cue signaled participants to stop; in the VW, the system automatically paused and disabled inputs. Correct answers earned one point toward the final SAGAT score. 




\subsubsection{Workload}

Task workload was assessed using the NASA-TLX \cite{hart_development_1988}. Participants rated six workload dimensions and performed pairwise comparisons to generate individual dimension weights. The overall workload score (0–100) was calculated as the weighted average of these ratings across all dimensions.

\subsubsection{Presence}

Presence was measured using the Slater-Usoh-Steed (SUS) presence questionnaire \cite{usoh_using_2000}. The SUS score was determined by evaluating participants' responses on a 7-point scale and counting the number of times a participant selected either six or seven as it indicates a high sense of presence. Conversely, we calculated a summed presence rating to focus on the magnitude of these high-tier experiences by averaging the values of all responses rated 6 or 7.

\subsection{Quantitative Analysis}

Normality was assessed using Shapiro–Wilk tests. Independent-samples t-tests were used when assumptions were met and Mann–Whitney U tests were conducted when normality was violated. For human factor measures repeated over trials (i.e., Trust, SA, Workload, and Presence), linear mixed-effects models (2x3) were employed with Condition as a between-subjects factor and trial as a within-subjects repeated factor to account for the nested data structure. Performance was analyzed using Trial 3 data to capture stabilized task execution. Additionally, regression analyses were conducted separately by condition to examine relationships between human factors and performance outcomes.

\subsection{Qualitative Data Collection and Analysis}

Qualitative methods were used for both data collection and analysis. During the trials, researchers engaged participants in a Think Aloud Protocol (TAP), encouraging them to verbalize their thoughts while performing tasks \cite{hu2017using}. Prompts included:
\begin{enumerate*}[label=(\arabic*)]
\item "What are you doing now?"
\item "Where are you?"
\item "Why did you do that?"
\item "Where are you heading?"
\item "What are you going to do next?"
\item "What made you decide to do that?"
\end{enumerate*}

After all three trials, participants completed semi-structured debriefing interviews focused on trust in the robot, performance strategies, and interaction experiences. Researchers also revisited notable TAP moments to clarify participants’ real-time reasoning.

All TAP sessions and interviews were audio and video recorded and transcribed. Two authors conducted a thematic analysis following Braun and Clarke’s six-phase framework \cite{clarke2017thematic}. Both researchers first familiarized themselves with the data and study context and then independently coded the dataset using an inductive approach.

Coders were aware of participants’ modality condition during coding, as the analysis aimed to examine differences between physical and virtual settings. Each participant’s full dataset was coded before moving to the next. Transcripts from the physical modality were coded first, followed by those from the virtual modality. Coding was iterative. Previously coded transcripts were revisited as new codes emerged, and patterns were examined both within and across modalities.

In later phases, the researchers met to review codes, discuss patterns, and refine developing themes. Codes were consolidated and reorganized through discussion. A visual mapping process using a Miro board \cite{chan2023miro} supported theme organization and comparison. During theme development, the researchers revisited relevant literature on concepts such as mental models of robots and mechanisms of trust in HRT. Meetings continued until consensus was reached and final themes were clearly defined.

\section{Results}
\subsection{Quantitative Results}

\subsubsection{Performance}
Performance was analyzed using Trial 3 to capture stabilized task execution after familiarization. Capture performance did not significantly differ between the RW ($M = 0.8793$, $SD = 0.0854$) and VW ($M = 0.9072$, $SD = 0.0639$) conditions, $t(26) = -0.98$, $p = .337$. Overheat performance also showed no significant difference between the RW ($Mean\ Rank = 13.5$) and VW ($Mean\ Rank = 15.5$), $U = 112.0$, $p = .541$. To examine the relationship between human factors and performance, regression analyses were conducted separately by condition. In the VW, SA was positively associated with capture performance ($\beta = 0.715$, $p = .016$) and overheat performance ($\beta = 0.683$, $p = .029$). In contrast, no significant associations were observed in the RW (capture: $\beta = -1.849$, $p = .098$; overheat: $\beta = 0.197$, $p = .569$). Trust, workload, and presence were not significantly associated with either performance metric in either condition.

\subsubsection{Trust} 
Across time, the linear mixed-effects model showed no significant main effect of Condition, $F(1, 92.35) = 0.20$, $p = .66$, no significant main effect of Trial Time, $F(3, 51.63) = 1.98$, $p = .13$, and no significant Condition × Trial Time interaction, $F(3, 51.63) = 0.07$, $p = .98$. At the final trial, trust did not significantly differ between the RW ($Mean\ Rank = 15.36$) and VW ($Mean\ Rank = 13.64$), $U = 86.0$, $p = .58$.

\subsubsection{Situational Awareness}
The linear mixed-effects model showed no significant main effect of Condition, $F(1, 26) = 1.58$, $p = .22$, no significant main effect of Trial Time, $F(2, 26) = 0.49$, $p = .62$, and no significant Condition × Trial Time interaction, $F(2, 26) = 3.29$, $p = .053$. At the final trial, SAGAT scores did not significantly differ between the RW ($Mean\ Rank = 14.18$) and VW ($Mean\ Rank = 14.82$), $U = 102.5$, $p = .839$.

\subsubsection{Workload}
The linear mixed-effects model showed no significant main effect of Condition, $F(1, 26) = 2.01$, $p = .168$, and no significant main effect of Trial, $F(2, 26) = 0.19$, $p = .829$. There was no significant Condition × Trial Time interaction, $F(2, 26) = 3.31$, $p = .052$. At the final trial, workload was higher in the RW ($M = 0.6960$, $SD = 0.1097$) than in the VW ($M = 0.5831$, $SD = 0.1695$), $t(26) = 2.09$, $p = .046$. Final trial sub-scale analyses showed that Mental Demand was higher in the VW ($Mean\ Rank = 18.82$) than in the RW ($Mean\ Rank = 10.18$), $U = 158.5$, $p = .004$, whereas Physical Demand was higher in the RW ($Mean\ Rank = 19.93$) than in the VW ($Mean\ Rank = 9.07$), $U = 22.0$, $p < .001$. Effort ($U = 58.0$, $p = .069$), Performance ($U = 98.5$, $p = 1.000$), Frustration ($U = 125.0$, $p = .227$), and Temporal Demand ($t(26) = 1.871$, $p = .073$) did not significantly differ between conditions.

\subsubsection{Presence}
The linear mixed-effects model showed no significant main effect of Condition, $F(1, 26) = 0.27$, $p = .61$, no significant main effect of Trial Time, $F(2, 26) = 3.06$, $p = .064$, and no significant Condition × Trial Time interaction, $F(2, 26) = 1.94$, $p = .16$. At the final trial, SUS scores did not significantly differ between the RW ($Mean\ Rank = 16.0$) and VW ($Mean\ Rank = 13.0$), $U = 77.0$, $p = .352$.

\subsection{Qualitative Results - Exploration of participant behavior in the HRT scenario}

\begin{table*}[t]

\centering
\vspace{6pt}
\caption{Detailed Themes for RW and VW Conditions}
\resizebox{\textwidth}{!}{%
\begin{tabular}{>{\raggedright\arraybackslash}p{3.5cm} 
                >{\raggedright\arraybackslash}p{6.5cm} 
                >{\raggedright\arraybackslash}p{6.5cm}}
\toprule
\textbf{Factor} & \textbf{RW Themes} & \textbf{VW Themes} \\
\midrule
General Trust & Expressed, but cautious & Expressed confidently \\
Sensitive Info Trust & Unwilling to fully trust robots with it & Willing to trust robots with it \\
Physical Workload & High (moving quickly between rooms) & Not applicable \\
Mental Workload & Lower (relied on MR screen) & Higher (continuous estimation and prediction) \\
Robot Perception & Described as flawed tools, like cameras or pagers & Described as assistants or extensions of self \\
Strategy & Stayed in MR most of the time & Actively moved or patrolled \\
Avatar Perception & N/A & Avatar speed \textit{frustrating}, impacted planning \\
\bottomrule
\end{tabular}
}
\label{tab:detailed_themes}
\end{table*}

Our thematic analysis of the TAP and post-scenario interviews identified patterns in how participants perceived and interacted with their robot teammates across conditions. Themes clustered around trust, workload, perception of the robot, and strategy. While some themes appeared in both modalities, their expression differed between them.

\subsubsection*{Trust}
Trust differed in how it was formed across conditions. In RW, participants based their trust on physical co-presence and direct observation. In the VW, trust was tied to alert reliability and system performance.

\textit{RW} participants expressed conditional trust linked to embodiment:

RW-P291: \textit{“During the trial, I said that there's a slight error and when they turned around…but I did trust the robots most of the time, because you know, they're in the room, physically.”}

However, this trust was not absolute. RW-P821 reflected that reliance was limited because they preferred verifying information themselves \textit{“I wanted to be able to see the hackers before I got the notifications to be able to react…”}

In the \textit{VW}, trust was articulated more confidently and tied directly to performance consistency:

VW-P447: \textit{“I did trust them to tell me that someone was going to be in there. There were no circumstances where I went to a room and found that it had been hacked, and I hadn't been alerted. And there were no instances where I got a false positive, or they said, there's someone in the room and I went there, and no one was there.”}

Similarly, VW-P522 stated \textit{“Yeah, based on the performance in this? I would.”}. Trust in the virtual condition was framed as outcome-based rather than presence-based.

\subsubsection*{Workload}

Participants in the different modalities also focused on different dimensions of workload. In the RW, participants emphasized physical strain and time pressure. In the VW, workload was described as cognitively demanding and strategy-driven.

Physical urgency and movement demands were central in the \textit{RW},:

RW-P666: \textit{“I'm thinking I can't take a breath...if I came back and there was a hacker in engineering center [Building B] I think I would have lost it."}

RW-P539 described time pressure and physical limitations \textit{“And I think it might not be possible to get to some of the rooms in time. So I might not walk as fast honestly.”}

Participants also used the monitoring room to regulate workload. RW-P717 stated \textit{“Nothing much, just get back to the control room and look at temperature and damage.”}. 
RW-P892 actively tracked temperature thresholds \textit{“They're about 45…they're about 50…I'm gonna wait till I hit about 70 just like I've been doing for the last ones...”}

In the \textit{VW}, physical effort decreased but cognitive demand increased. VW-P916 stated \textit{“It was mentally demanding the task itself”}. VW-P944 demonstrated anticipatory reasoning \textit{“If there's a 30 degree difference between the two if one reaches 80, the other will be at 50”}. 

Strategic uncertainty further increased mental load \textit{“I think I gonna have to just guess at where I think it's gonna get hit next and head there”}. 
VW-P481 added \textit{“I don't really know what my plan is anymore… I don't really know what to do anymore”}. In the virtual condition, workload was framed as prediction, planning, and continuous recalculation.

\subsubsection*{Perception of Robot}

Perception of the role of the robots and their qualities differed across modalities. In the RW, robots were often described as limited tools. In the VW, they were more frequently framed as assistants or teammates.

In the \textit{RW}, participants minimized the robot’s autonomy. RW-P821 described them as \textit{“…more like glorified security cameras...”}. RW-P539 referred to them as \textit{“...pagers”}. 

Concerns about reliability and visibility were also present. RW-P666 stated \textit{“I think I get anxious every time the cameras turn around”}. RW-P580 added \textit{“I was like afraid… you wouldn't be able to see the hacker from the first go”}.

In the \textit{VW}, participants described the robot as supportive and task-oriented. VW-P916 stated \textit{“The role of the robot was an assistant to help me accomplish my task”}.

VW-P267 emphasized performance \textit{“They did a good job. Did a great job at it”}. VW-P849 framed the robot as perceptual extension
\textit{“…it's kind of like your eyes and your feet there”}.

Even criticism reflected relational framing. VW-P610 stated \textit{“It [robot] just didn't see the guy in time… that's a pretty bad teammate...”}. In the VW, the robot was evaluated as a collaborator rather than a device.

\subsubsection*{Strategy}

Strategic behavior differed in spatial positioning and reliance on robot input.

In the \textit{RW}, participants predominantly adopted a centralized “home base” strategy in the monitoring room. RW-P750 stated \textit{“Okay, yeah now I'm heading back to the monitor room ”}. 

RW-P821 justified this choice \textit{“Here… you have the most control during the trial”}. RW-P717 described monitoring behavior:
\textit{“just watching the monitors… monitoring everything on the monitors”}.

Participants also adjusted reliance on robot alerts. RW-P291 reflected RW-P291: \textit{“I think in the beginning, I was relying on them too much… then later… I have to make decisions, not just based on robots…”}

In the \textit{VW}, participants more frequently described patrol-based strategies. VW-P447 stated \textit{“I'm probably just gonna go float back and forth between the two different server rooms”}.

VW-P610 explained \textit{“If I just sat in the server room, I would be too slow… I'd rather patrol around”}. VW-P522 added \textit{“See if I can be out walking before the intruder is alert ”}.

These excerpts show a shift toward anticipatory movement and spatial coverage in the virtual condition.

\subsubsection*{Avatar Perception}

Avatar perception was unique to the VW. Participants frequently commented on movement speed and control responsiveness. 
VW-P849 stated \textit{“But my pace is so so slow ”}. VW-P481 similarly noted \textit{“I'm like walking really slow from what it feels like ”}. VW-P172 explained:

VW-P172: \textit{“It definitely took some getting used to… movement speed. That was atrocious to deal with… especially in a task where you have to be at a certain place before something happens.”}

No comparable embodiment-related concerns were expressed in the RW condition.

\section{Discussion}

The goal of this study was to examine whether findings from VW and RW HRT user studies are comparable by investigating differences in performance, human factors, and behavior across modalities. Our quantitative results showed no significant differences in performance, trust, or SA between VW and RW. However, the qualitative findings and workload patterns suggest that participants engaged with the task differently across conditions. Although overall outcomes were comparable, participants described different strategies, different sources of demand, and different ways of thinking about the robots in the VW and RW. These findings indicate that while measured outcomes were similar, the experience and approach to teaming varied by modality.

\subsection{Modality Shifting Trust}
Our quantitative analysis supported H1, revealing no significant difference in trust scores between VW and RW. These results align with Plomin et al. \cite{plomin2023virtual}, confirming that perceived trust remains comparable across both modalities. In contrast, qualitative findings suggest differences in how participants constructed that trust. In the RW, trust was often grounded in physical co-presence. Participants described the robots as entities sharing their environment and frequently adopted a trust but verify approach, preferring to visually confirm intrusions even when alerts were provided. Alternatively, participants in the VW framed trust in terms of system reliability and alert consistency. Robots were described more as dependable sensing agents, and trust was tied to whether notifications were accurate and timely.

\subsection{The Workload Split}
The results for H2 highlight differences in how workload was distributed across modalities. Overall workload was higher in the RW at the final trial ($p = .046$). Subscale analyses revealed a clear divergence: RW participants reported higher Physical Demand, whereas VW participants reported significantly higher Mental Demand. Qualitative findings help contextualize this pattern. RW participants described urgency in moving between rooms and often positioned themselves in a central location to manage time pressure and locomotion demands. Conversely, VW participants emphasized mental demand in regards to prediction, planning, and continuous interpretation of system information. Several also noted difficulty with avatar movement speed and spatial coordination, possibly contributing to VW's higher mental demand.

This distribution aligns with prior findings suggesting that virtual environments can increase cognitive demand \cite{sakib2021physiological}. Rather than indicating a simple increase or decrease in workload, our results suggest that modality shifts where effort is concentrated, from embodied movement in the RW to cognitive coordination in the VW.

\subsection{SA as a Performance Driver in Simulation}
Although SAGAT scores did not differ between the VW and RW, the relationship between SA and performance varied by condition. In the VW, SA significantly predicted capture and overheat performance. In the RW, no significant association between SA and performance was observed. It's possible differences in strategy present between the two modalities affected this outcome. In the VW, participants utilized more patrol-based strategies which required participants to use SA to predict where problems would occur and decide where to move next. While, in the RW, the centralized “home base” strategy allowed participants to monitor the system from one location, meaning performance depended more on the time needed to physically reach each room. These results suggest that comparable SA scores across modalities do not necessarily translate into comparable performance relationships. Consistent with Horsch et al. \cite{horsch2013comparing}, overall SA levels may appear similar when task conditions are controlled. However, our findings indicate that the role SA plays in supporting task performance may differ by modality.

\subsection{Presence and Task Focus}
Contrary to H4, presence did not differ between the VW and RW. This contrasts prior findings suggesting stronger co-presence in RW settings \cite{li2019comparing, weistroffer2014assessing}, while aligning with work showing that immersive virtual environments can approximate RW presence under certain conditions \cite{plomin2023virtual}.

Despite comparable presence ratings, participants' strategies and robot framing still diverged by modality. In addition, several VW participants commented on avatar movement speed and control responsiveness, noting that these factors influenced how they navigated and planned within the environment. These differences suggest that similar levels of reported presence do not necessarily translate to similar interaction patterns in HRT tasks.

\section{Conclusion}

This paper investigated the validity of VW simulations as a proxy for studying RW HRT through a mixed-methods security scenario study. By examining performance, human factors, and behavioral strategies across both modalities, we provide a nuanced map of where virtual and physical environments align and where they diverge. While quantitative results showed no significant differences in performance, trust, or SA, the qualitative data reveals that participants achieved identical outcomes through fundamentally different thought processes. Specifically in the RW, trust is grounded in physical co-presence and manual verification. While in the VW, it shifts toward system reliability and data consistency. For workload, modality does not change the amount of effort, but its nature. RW demand is primarily physical, whereas VW demand is cognitive. Regarding SA, it significantly predicted performance in the VW, but not in the RW, suggesting that virtual operators rely more heavily on mental mapping to compensate for the lack of physical cues. Furthermore, while reported presence was statistically comparable, it did not translate into identical behavioral strategies: RW participants tended to centralize themselves spatially and view robots as tools, whereas VW participants adopted active patrol strategies and viewed robots as assistants. Ultimately, VW simulations serve as a high-fidelity proxy for approximating performance and general human factors; however, these end outcomes may be a consequence of differing behavioral strategies and thought processes exhibited by users due to interaction qualities unique to each modality.

\section{Acknowledgment}

The authors wish to acknowledge the technical and financial support of the Automotive Research Center (ARC) in accordance with the Cooperative Agreement W56HZV-24-2-0001 US Army DEVCOM Ground Vehicle Systems Center (GVSC) Warren, MI.

\bibliography{RCG}

@inproceedings{li2019comparing,
  title={Comparing human-robot proxemics between virtual reality and the real world},
  author={Li, Rui and van Almkerk, Marc and van Waveren, Sanne and Carter, Elizabeth and Leite, Iolanda},
  booktitle={2019 14th ACM/IEEE international conference on human-robot interaction (HRI)},
  pages={431--439},
  year={2019},
  organization={IEEE}
}

@inproceedings{wijnen2020performing,
  title={Performing human-robot interaction user studies in virtual reality},
  author={Wijnen, Luc and Bremner, Paul and Lemaignan, S{\'e}verin and Giuliani, Manuel},
  booktitle={2020 29th IEEE international conference on robot and human interactive communication (RO-MAN)},
  pages={794--794},
  year={2020},
  organization={IEEE}
}

@inproceedings{weistroffer2014assessing,
  title={Assessing the acceptability of human-robot co-presence on assembly lines: A comparison between actual situations and their virtual reality counterparts},
  author={Weistroffer, Vincent and Paljic, Alexis and Fuchs, Philippe and Hugues, Olivier and Chodacki, Jean-Paul and Ligot, Pascal and Morais, Alexandre},
  booktitle={The 23rd IEEE International Symposium on Robot and Human Interactive Communication},
  pages={377--384},
  year={2014},
  organization={IEEE}
}

@inproceedings{endsley_situation_1988,
	location = {Dayton, {OH}, {USA}},
	title = {Situation awareness global assessment technique ({SAGAT})},
	url = {http://ieeexplore.ieee.org/document/195097/},
	doi = {10.1109/NAECON.1988.195097},
	eventtitle = {{IEEE} 1988 National Aerospace and Electronics Conference},
	pages = {789--795},
	booktitle = {Proceedings of the {IEEE} 1988 National Aerospace and Electronics Conference},
	publisher = {{IEEE}},
	author = {Endsley, M.R.},
	urldate = {2022-10-04},
	date = {1988},
}

@incollection{mittu_measuring_2016,
	location = {Boston, {MA}},
	title = {Measuring Trust in Human Robot Interactions: Development of the “Trust Perception Scale-{HRI}”},
	isbn = {978-1-4899-7666-6 978-1-4899-7668-0},
	url = {http://link.springer.com/10.1007/978-1-4899-7668-0_10},
	shorttitle = {Measuring Trust in Human Robot Interactions},
	pages = {191--218},
	booktitle = {Robust Intelligence and Trust in Autonomous Systems},
	publisher = {Springer {US}},
	author = {Schaefer, Kristin E.},
	editor = {Mittu, Ranjeev and Sofge, Donald and Wagner, Alan and Lawless, W.F.},
	urldate = {2022-09-27},
	date = {2016},
	langid = {english},
	doi = {10.1007/978-1-4899-7668-0_10},
}

@incollection{hart_development_1988,
	title = {Development of {NASA}-{TLX} (Task Load Index): Results of Empirical and Theoretical Research},
	volume = {52},
	isbn = {978-0-444-70388-0},
	url = {https://linkinghub.elsevier.com/retrieve/pii/S0166411508623869},
	shorttitle = {Development of {NASA}-{TLX} (Task Load Index)},
	pages = {139--183},
	booktitle = {Advances in Psychology},
	publisher = {Elsevier},
	author = {Hart, Sandra G. and Staveland, Lowell E.},
	urldate = {2023-03-28},
	date = {1988},
	langid = {english},
	doi = {10.1016/S0166-4115(08)62386-9},
}

@article{usoh_using_2000,
	title = {Using Presence Questionnaires in Reality},
	volume = {9},
	issn = {1054-7460},
	url = {https://direct.mit.edu/pvar/article/9/5/497-503/18368},
	doi = {10.1162/105474600566989},
	pages = {497--503},
	number = {5},
	journaltitle = {Presence: Teleoperators and Virtual Environments},
	shortjournal = {Presence: Teleoperators \& Virtual Environments},
	author = {Usoh, Martin and Catena, Ernest and Arman, Sima and Slater, Mel},
	urldate = {2023-03-28},
	date = {2000-10},
	langid = {english},
}

@article{hu2017using,
  title={Using think-aloud protocol in self-regulated reading research},
  author={Hu, Jingjing and Gao, Xuesong Andy},
  journal={Educational Research Review},
  volume={22},
  pages={181--193},
  year={2017},
  publisher={Elsevier}
}

@article{clarke2017thematic,
  title={Thematic analysis},
  author={Clarke, Victoria and Braun, Virginia},
  journal={The journal of positive psychology},
  volume={12},
  number={3},
  pages={297--298},
  year={2017},
  publisher={Taylor \& Francis}
}

@article{guetterman2015integrating,
  title={Integrating quantitative and qualitative results in health science mixed methods research through joint displays},
  author={Guetterman, Timothy C and Fetters, Michael D and Creswell, John W},
  journal={The Annals of Family Medicine},
  volume={13},
  number={6},
  pages={554--561},
  year={2015},
  publisher={Annals Family Med}
}

@book{creswell2017designing,
  title={Designing and conducting mixed methods research},
  author={Creswell, John W and Clark, Vicki L Plano},
  year={2017},
  publisher={Sage publications}
}

@article{chan2023miro,
  title={Miro: Promoting collaboration through online whiteboard interaction},
  author={Chan, Thomas Anthony Chun Hun and Ho, Jason Man-Bo and Tom, Michael},
  journal={RELC Journal},
  pages={00336882231165061},
  year={2023},
  publisher={SAGE Publications Sage UK: London, England}
}

@article{robinson2024human,
  title={Human-Robot Team Performance Compared to Full Robot Autonomy in 16 Real-World Search and Rescue Missions: Adaptation of the DARPA Subterranean Challenge},
  author={Robinson, Nicole and Williams, Jason and Howard, David and Tidd, Brendan and Talbot, Fletcher and Wood, Brett and Pitt, Alex and Kottege, Navinda and Kuli{\'c}, Dana},
  journal={ACM Transactions on Human-Robot Interaction},
  volume={14},
  number={1},
  pages={1--30},
  year={2024},
  publisher={ACM New York, NY}
}

@inproceedings{wang2015intelligent,
  title={Intelligent agents for virtual simulation of human-robot interaction},
  author={Wang, Ning and Pynadath, David V and Unnikrishnan, KV and Shankar, Santosh and Merchant, Chirag},
  booktitle={Virtual, Augmented and Mixed Reality: 7th International Conference, VAMR 2015, Held as Part of HCI International 2015, Los Angeles, CA, USA, August 2-7, 2015, Proceedings 7},
  pages={228--239},
  year={2015},
  organization={Springer}
}

@inproceedings{gervits2020toward,
  title={Toward Genuine Robot Teammates: Improving Human-Robot Team Performance Using Robot Shared Mental Models.},
  author={Gervits, Felix and Thurston, Dean and Thielstrom, Ravenna and Fong, Terry and Pham, Quinn and Scheutz, Matthias},
  booktitle={Aamas},
  pages={429--437},
  year={2020}
}

@inproceedings{walker2024cyber,
  title={The cyber-physical control room: A mixed reality interface for mobile robot teleoperation and human-robot teaming},
  author={Walker, Michael E and Gramopadhye, Maitrey and Ikeda, Bryce and Burns, Jack and Szafir, Daniel},
  booktitle={Proceedings of the 2024 ACM/IEEE International Conference on Human-Robot Interaction},
  pages={762--771},
  year={2024}
}

@article{mcneese2018teaming,
  title={Teaming with a synthetic teammate: Insights into human-autonomy teaming},
  author={McNeese, Nathan J and Demir, Mustafa and Cooke, Nancy J and Myers, Christopher},
  journal={Human factors},
  volume={60},
  number={2},
  pages={262--273},
  year={2018},
  publisher={Sage Publications Sage CA: Los Angeles, CA}
}

@article{hopko2022human,
  title={Human factors considerations and metrics in shared space human-robot collaboration: A systematic review},
  author={Hopko, Sarah and Wang, Jingkun and Mehta, Ranjana},
  journal={Frontiers in Robotics and AI},
  volume={9},
  pages={799522},
  year={2022},
  publisher={Frontiers Media SA}
}

@misc{turtlebot2,
  author       = {{Open Source Robotics Foundation}},
  title        = {{TurtleBot 2: Open-Source Personal Robot Kit}},
  year         = {2013},
  howpublished = {\url{http://www.turtlebot.com}},
  note         = {Accessed: 2026-01-08}
}

@article{plomin2023virtual,
  title={Virtual reality check: a comparison of virtual reality, screen-based, and real world settings as research methods for HRI},
  author={Plomin, Jana and Schweidler, Paul and Oehme, Astrid},
  journal={Frontiers in Robotics and AI},
  volume={10},
  pages={1156715},
  year={2023},
  publisher={Frontiers Media SA}
}

@article{wolf2023and,
  title={How and when can robots be team members? Three decades of research on human--robot teams},
  author={Wolf, Franziska Doris and Stock-Homburg, Ruth Maria},
  journal={Group \& Organization Management},
  volume={48},
  number={6},
  pages={1666--1744},
  year={2023},
  publisher={Sage Publications Sage CA: Los Angeles, CA}
}

@article{adami2022impact,
  title={Impact of VR-based training on human--robot interaction for remote operating construction robots},
  author={Adami, Pooya and Rodrigues, Patrick B and Woods, Peter J and Becerik-Gerber, Burcin and Soibelman, Lucio and Copur-Gencturk, Yasemin and Lucas, Gale},
  journal={Journal of Computing in Civil Engineering},
  volume={36},
  number={3},
  pages={04022006},
  year={2022},
  publisher={American Society of Civil Engineers}
}

@article{abdulazeem2023human,
  title={Human factors considerations for quantifiable human states in physical human--robot interaction: a literature review},
  author={Abdulazeem, Nourhan and Hu, Yue},
  journal={Sensors},
  volume={23},
  number={17},
  pages={7381},
  year={2023},
  publisher={MDPI}
}

@inproceedings{horsch2013comparing,
  title={Comparing performance and situation awareness in USAR unit tasks in a virtual and real environment},
  author={Horsch, Corine HG and Smets, Nanja JJM and Neerincx, Mark A and Cuijpers, Raymond H},
  booktitle={10th International Conference on Information Systems for Crisis Response and Management 2013 (ISCRAM 2013), 12-15 May 2013, Baden Baden, Germany},
  pages={ID144},
  year={2013}
}

@article{sakib2021physiological,
  title={Physiological data models to understand the effectiveness of drone operation training in immersive virtual reality},
  author={Sakib, Md Nazmus and Chaspari, Theodora and Behzadan, Amir H},
  journal={Journal of Computing in Civil Engineering},
  volume={35},
  number={1},
  pages={04020053},
  year={2021},
  publisher={American Society of Civil Engineers}
}

@article{tsoi2024influence,
  title={Influence of simulation and interactivity on human perceptions of a robot during navigation tasks},
  author={Tsoi, Nathan and Sterneck, Rachel and Zhao, Xuan and V{\'a}zquez, Marynel},
  journal={ACM Transactions on Human-Robot Interaction},
  volume={13},
  number={4},
  pages={1--19},
  year={2024},
  publisher={ACM New York, NY}
}

@book{campbell2015experimental,
  title={Experimental and quasi-experimental designs for research},
  author={Campbell, Donald T and Stanley, Julian C},
  year={2015},
  publisher={Ravenio books}
}
\end{document}